\documentclass{article}

\usepackage{PRIMEarxiv}

\usepackage[utf8]{inputenc} 
\usepackage[T1]{fontenc}    
\usepackage[hidelinks]{hyperref}       
\usepackage{url}            
\usepackage{booktabs}       
\usepackage{amsfonts}       
\usepackage{nicefrac}       
\usepackage{microtype}      
\usepackage{lipsum}
\usepackage{fancyhdr}       
\usepackage{graphicx}       
\usepackage[round,authoryear]{natbib}

\usepackage{amsmath}
\usepackage{amssymb}
\usepackage{float}
\usepackage{tikz}
\usepackage{subcaption}
\usepackage{array}
\usepackage{xcolor}
\usepackage{booktabs}
\usepackage{multirow}
\usepackage{siunitx}

\usepackage{tikz}
\usetikzlibrary{calc, shapes.geometric, arrows.meta, positioning}

\title{Spatio-Temporal Detection of Photovoltaic Systems on Industrial Building Rooftops Using Satellite Imagery and Machine Learning}

\author{
    \begin{tabular}{p{0.85\textwidth}}
        \centering
        Frederick Bruch$^{1,2}$, Christian Knoth$^{1}$, Florian Detsch$^{3}$, \\
        Edzer Pebesma$^{1}$, Hanna Meyer$^{2}$ \\[0.2cm]
        \small \textbf{1} University of Münster, Institute for Geoinformatics, Heisenbergstraße 2, 48149 Münster, Germany \\
        \small \textbf{2} University of Münster, Institute of Landscape Ecology, Heisenbergstraße 2, 48149 Münster, Germany \\
        \small \textbf{3} Elgin Energy GmbH, Hafenbad 22, 8907 Ulm, Germany
    \end{tabular}
}

\begin{document}
\maketitle

\begin{abstract}
The global transition toward carbon-neutral energy systems is heavily dependent on the rapid deployment of solar photovoltaic (PV) infrastructure, yet the speed of this expansion has outpaced the ability of administrative registries to maintain accurate records. While high-resolution commercial imagery is often used for monitoring, its high cost and infrequent updates can create an information gap for grid operators, energy infrastructure developers and policymakers. This study addresses the lack of scalable monitoring for industrial-scale rooftop PV systems by developing a machine learning framework that utilizes open-access, multispectral Sentinel-2 satellite imagery.
By integrating building geometries from OpenStreetMap with Sentinel-2 reflectance data, a Random Forest model was trained to identify the specific spectral signatures of PV modules. Moving beyond theoretical modeling, this work emphasizes practical implementation through the development
of an interactive web-based prototype for on-demand site monitoring. The application enables users to retrieve up-to-date satellite observations and assess the presence of PV systems on industrial building rooftops. The tool incorporates quality metrics that inform users whether a prediction is supported by sufficient training data density or requires manual verification. This end-to-end framework demonstrates that open-source medium-resolution data can effectively bridge this monitoring gap.
\end{abstract}


\section{Introduction}
\label{chap:introduction}

The rapid expansion of solar photovoltaic (PV) infrastructure plays a central role in efforts to meet reduction targets for greenhouse gas emissions across all administrative levels.
Companies that specialize in selling or leasing PV systems for industrial rooftops are major contributors to this expansion. To sustain this progress, precise and timely monitoring of existing PV infrastructure as well as the identification of suitable locations for future installations is needed. However, the rapid deployment of PV infrastructure has outpaced the capabilities of administrative registries to track these assets accurately \citep{Stowell2020}. Although some countries maintain registries—such as Germany’s official energy registry, the Marktstammdatenregister (MaStR)—the information they contain is often outdated. This creates an information gap for grid operators, energy policymakers, and environmental researchers.

\begin{figure}[htbp]
    \centering
    \begin{subfigure}[b]{0.32\textwidth}
        \centering
        \includegraphics[width=\textwidth]{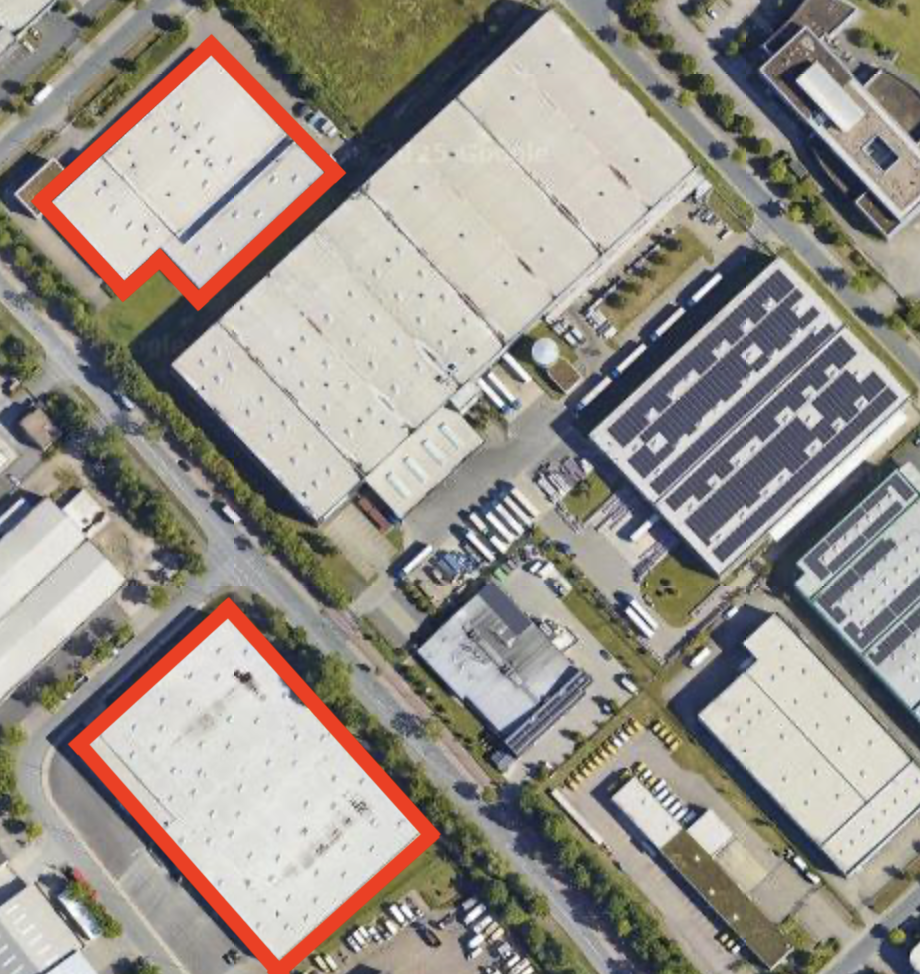}
        \caption{High-Resolution (Google Maps)}
        \label{fig:google_example}
    \end{subfigure}
    \hfill
    \begin{subfigure}[b]{0.32\textwidth}
        \centering
        \includegraphics[width=\textwidth]{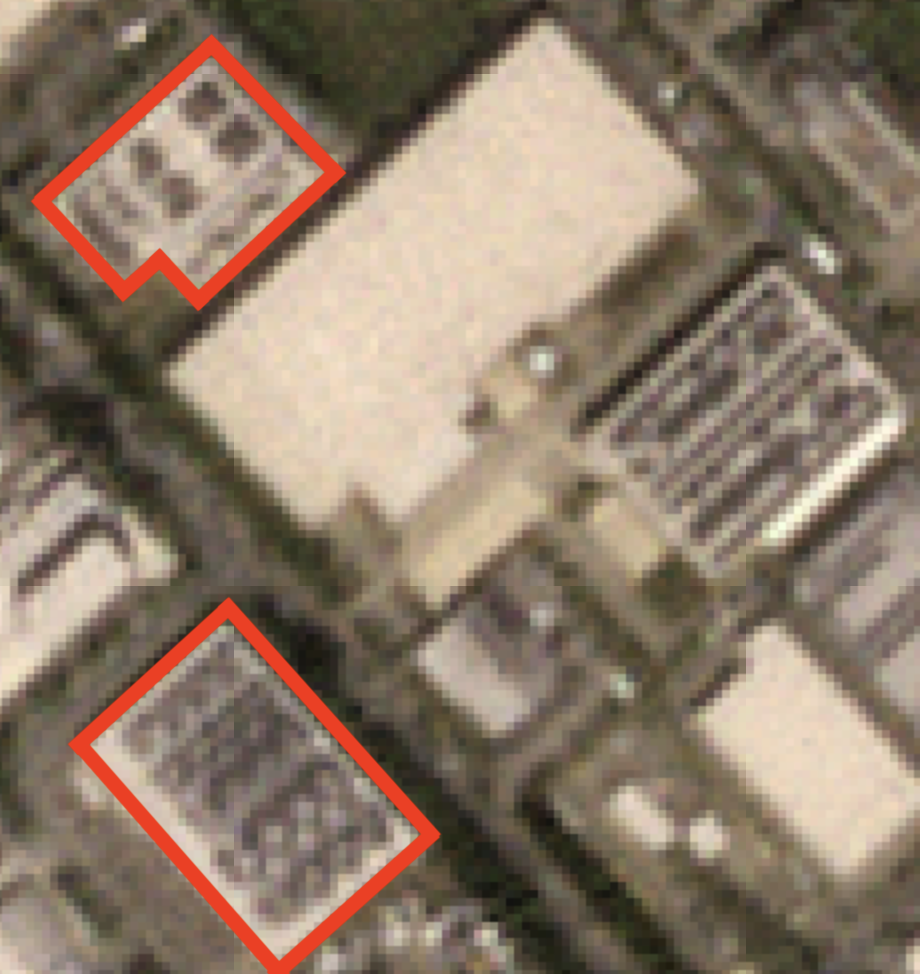}
        \caption{High-Resolution (Planet Labs)}
        \label{fig:planet_example}
    \end{subfigure}
    \hfill
    \begin{subfigure}[b]{0.32\textwidth}
        \centering
        \includegraphics[width=\textwidth]{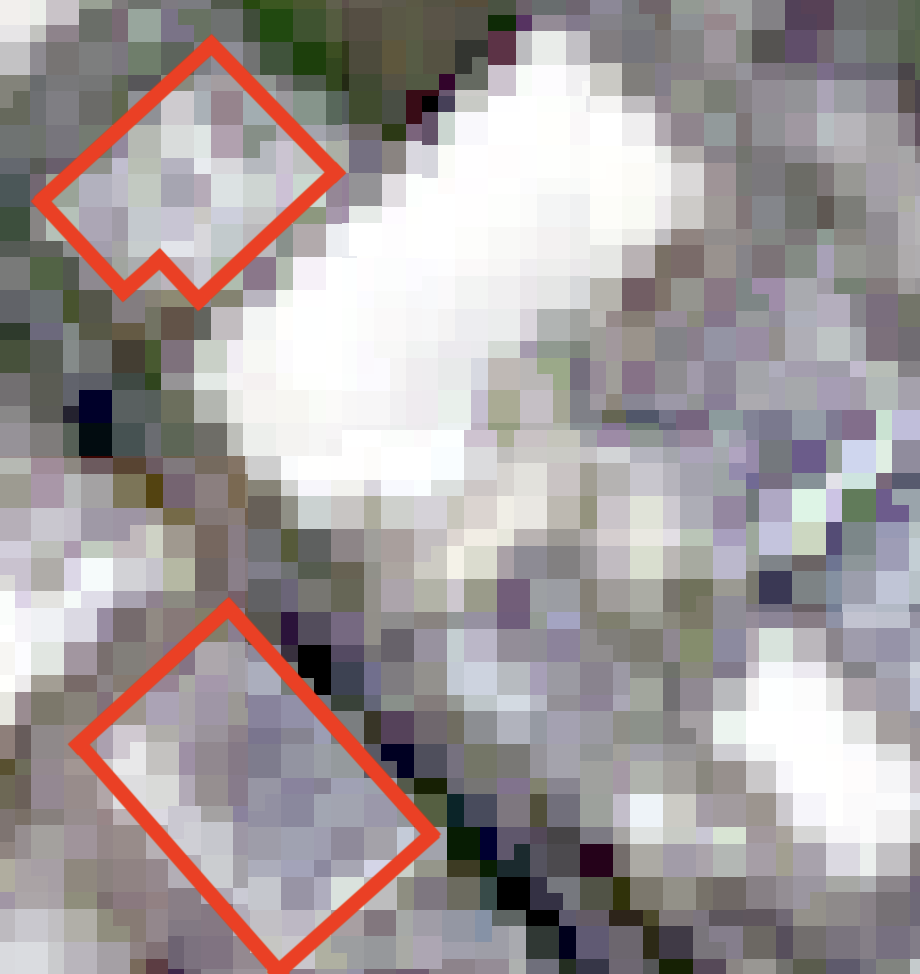}
        \caption{Medium-Resolution (Sentinel-2)}
        \label{fig:sentinel_example}
    \end{subfigure}
    \caption{The temporal and spatial trade-off in satellite monitoring for PV installations. (\subref{fig:google_example}) Outdated high-resolution imagery shows empty rooftops. (\subref{fig:planet_example}) Expensive commercial high-resolution imagery confirms a new PV system. (\subref{fig:sentinel_example}) Sentinel-2 medium-resolution imagery provides up-to-date information but does not clearly differentiate the PV system from the roof in the true-color composite compared to the higher resolution data.}
    \label{fig:temporal_spatial_tradeoff}
\end{figure}

Remote sensing has a high potential for systematic and large-scale detection, as it allows for repeated observations of the Earth's surface.  Several studies have already shown that remote sensing data allow for reliable detection of PV systems \citep[e.g.][]{Pena2024, kleebauer_multi-resolution_2023}. Most studies relied on commercial high-resolution satellite or aerial imagery, which allows for the detection of even small PV systems. This, however, typically comes with a practical trade-off: affordable and workable options are updated infrequently, while sources that offer both high spatial and high temporal resolution, such as the PlanetScope constellation, are only available at substantial cost if used beyond research purposes. As a result, developers often rely on imagery that has high spatial resolution but is outdated, introducing inefficiency and avoidable errors when searching for potential sites. This is especially problematic if developers with insufficient awareness do not realize the risk that the aerial imagery built into their preferred off-the-shelf solution (e.g. Esri WorldImagery, Mapbox and the like) could be outdated. This trade-off and its potential practical implications are demonstrated in \autoref{fig:temporal_spatial_tradeoff}.

The European Union's Copernicus Sentinel-2 mission \citep{Drusch2012} provides open and frequently collected multispectral satellite imagery (visible, near infrared, and shortwave infrared (SWIR)) at 10 m or 20 m spatial resolution. With the Sentinel-2 constellation, the satellite revisit time is five days at the equator~\citep{copernicus_sentinel-2_2025}. This regular acquisition enables the temporal consistency needed to monitor rapid changes. With its medium spatial resolution and acquisition frequency, this imagery provides a practical alternative to costly aerial surveys or high-resolution satellite data when regularly updated information is needed. Sentinel-2 data have been successfully used for detection of utility-scale PV installations \citep{kruitwagen_global_2021, li_global_2025, dalagan_spatial_2024}, however, the potential for detecting PV systems on the scale of industrial rooftops and in complex urban environments has not yet been studied.

To derive information on PV systems from remote sensing imagery, the application of machine learning has created effective solutions for solar monitoring at both utility and residential scales, yet a significant gap remains for industrial-scale infrastructure.
At a few centimeter resolution as provided by aerial images, deep learning approaches such as convolutional neural networks (CNN) have shown high potential due to the ability to learn the unique visual patterns of PV systems \citep[e.g.][]{Pena2024}. However, when lower resolution satellite data are being used (i.e. to benefit from the higher temporal resolution), such approaches might not be suitable since spatial patterns of PV systems are unlikely to be resolved at 10 m spatial resolution, unless massive utility-scale PV systems are to be detected (\citet{kruitwagen_global_2021, li_global_2025, dalagan_spatial_2024}).
Approaches that learn from the spectral properties of PV systems rather than from spatial patterns are an alternative in such cases. \citet{feng202410} used Random Forest on multispectral Sentinel-2 imagery to create a 10 meter national PV map of ground-mounted PV power stations in China, and \citet{zhang_mapping_2023} combined Sentinel-2 with Sentinel-1 data to classify large-scale PV installations on land and water. A Random Forest approach on segmented Sentinel-1 and Sentinel-2 data by \citet{plakman2022solar} detected solar parks with high accuracy. These tree-based models focus on the panels’ spectral signature rather than shape or textural patterns (unlike CNNs on high-resolution images), and have proven effective in large-area PV monitoring \citep{shimada_satellite-based_2025}. However, the application of these methods to industrial rooftops remains unexplored. Unlike ground-mounted parks, which offer relatively homogeneous spectral signatures, rooftop installations are often embedded within complex urban environments where the surrounding infrastructure creates spectral noise, making accurate detection significantly more challenging.

This work aims to explore the potential of using Sentinel-2 imagery for the detection of PV systems on industrial rooftops. Building on these findings, we develop and validate an open-data-based, scalable framework for automatic detection of PV installations on rooftops of industrial buildings for entire Germany. 
The resulting application can be operationally used by stakeholders requiring dynamic tools that allow for for monitoring and planning of PV infrastructure.
To accomplish these objectives, we utilize OpenStreetMap (OSM) to generate a training dataset of rooftop PV systems and various other rooftop structures without PV systems. By combining the geometries from OSM with Sentinel-2 multispectral observations, a Random Forest model is trained to recognize the specific spectral properties of solar infrastructure. The model is applied to the entire area of Germany to predict PV systems on industrial rooftops with associated uncertainty measures \citep[i.e., the Local Point Density][]{schumacher_estimation_2024}.
To demonstrate the scalability and practical applicability of the framework, we develop a prototypical open-source web application, which allows stakeholders to interactively apply the model to up-to-date imagery for on-demand site monitoring.

\section{Materials and Methods}
\label{chap:methodology}

The objective of this work is to develop an automated framework for detecting and visualising industrial rooftop PV systems using freely available medium-resolution satellite imagery. We applied this approach across Germany to evaluate the framework's performance over a large geographic extent. The following sections detail the data sources, the logic of the sampling pipeline, model training and reliability assessment.

\subsection{Data sources and preprocessing} 
\label{subsec:data_sources} 

\begin{figure}[ht!]
    \centering
    \includegraphics[width=0.75\textwidth]{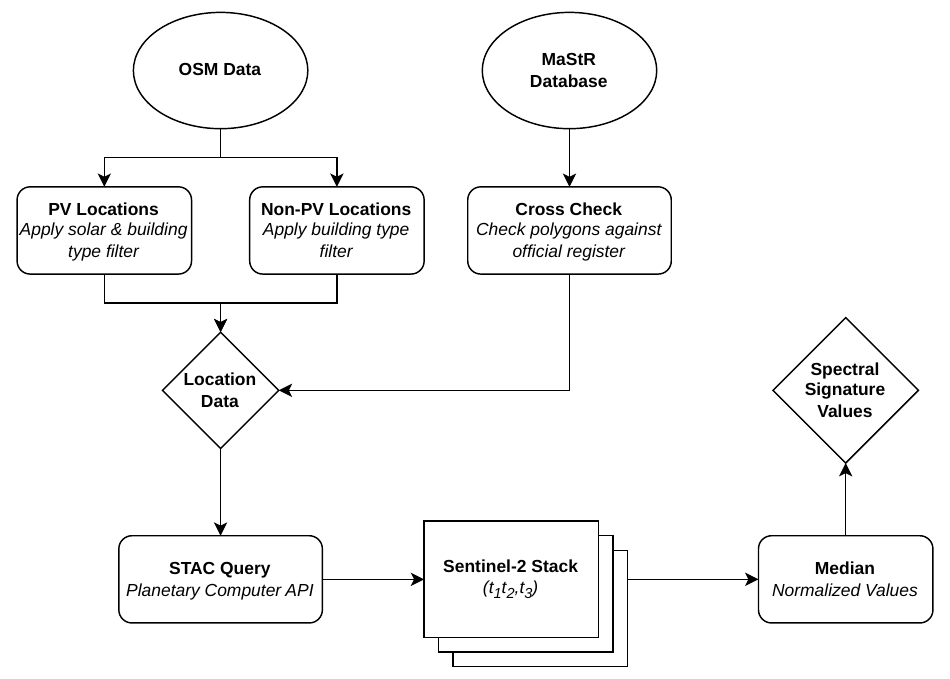}
    \caption{Methodology pipeline. From OSM data to spectral signature values.}
    \label{fig:unified_workflow}
\end{figure}

\begin{figure}[htbp]
    \centering
    \begin{subfigure}[b]{0.49\textwidth}
        \centering
        \includegraphics[width=\textwidth]{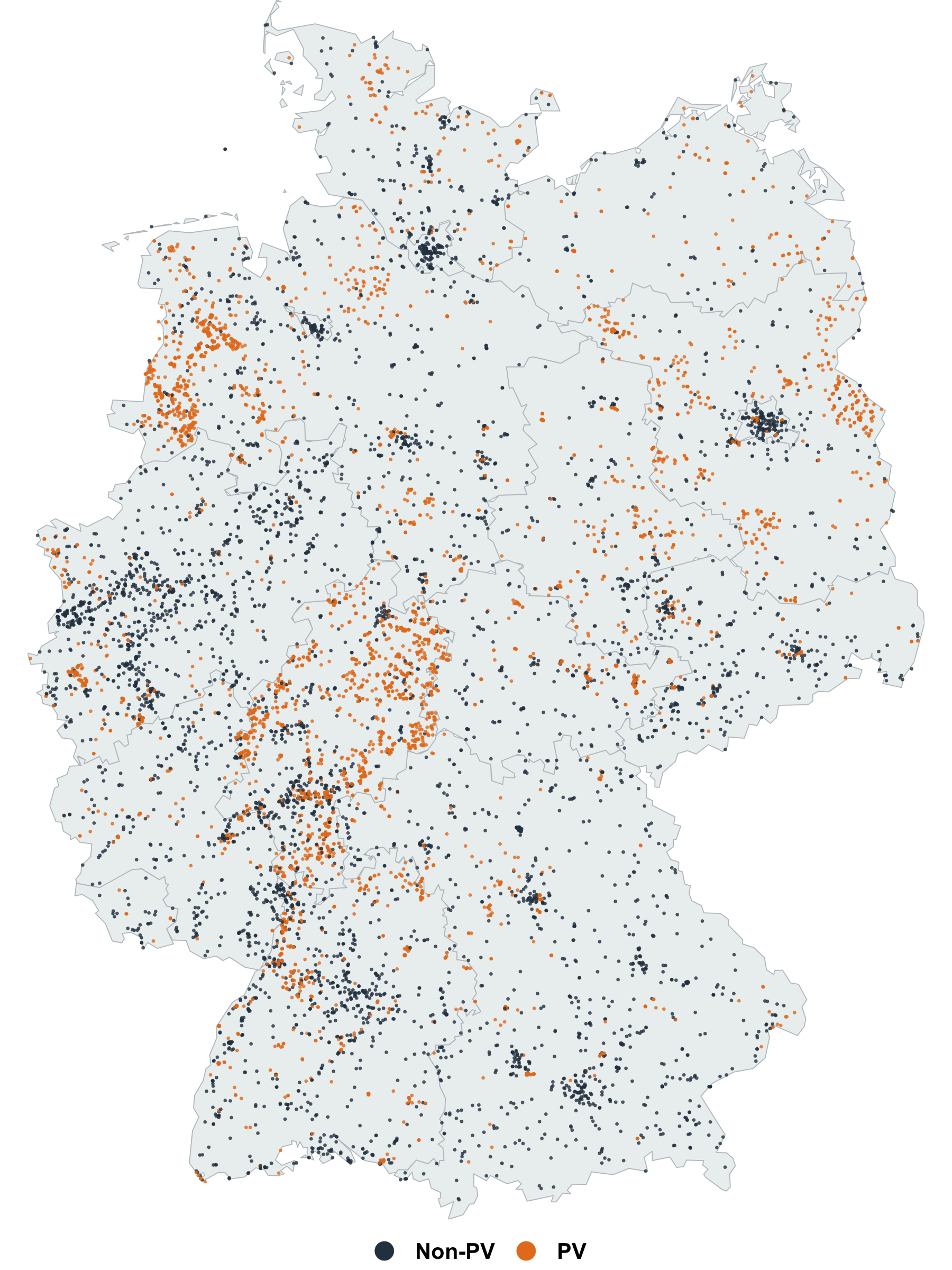}
        \caption{Sample Distribution}
        \label{fig:data_overview_samples}
    \end{subfigure}
    \hfill
    \begin{subfigure}[b]{0.49\textwidth}
        \centering
        \includegraphics[width=\textwidth]{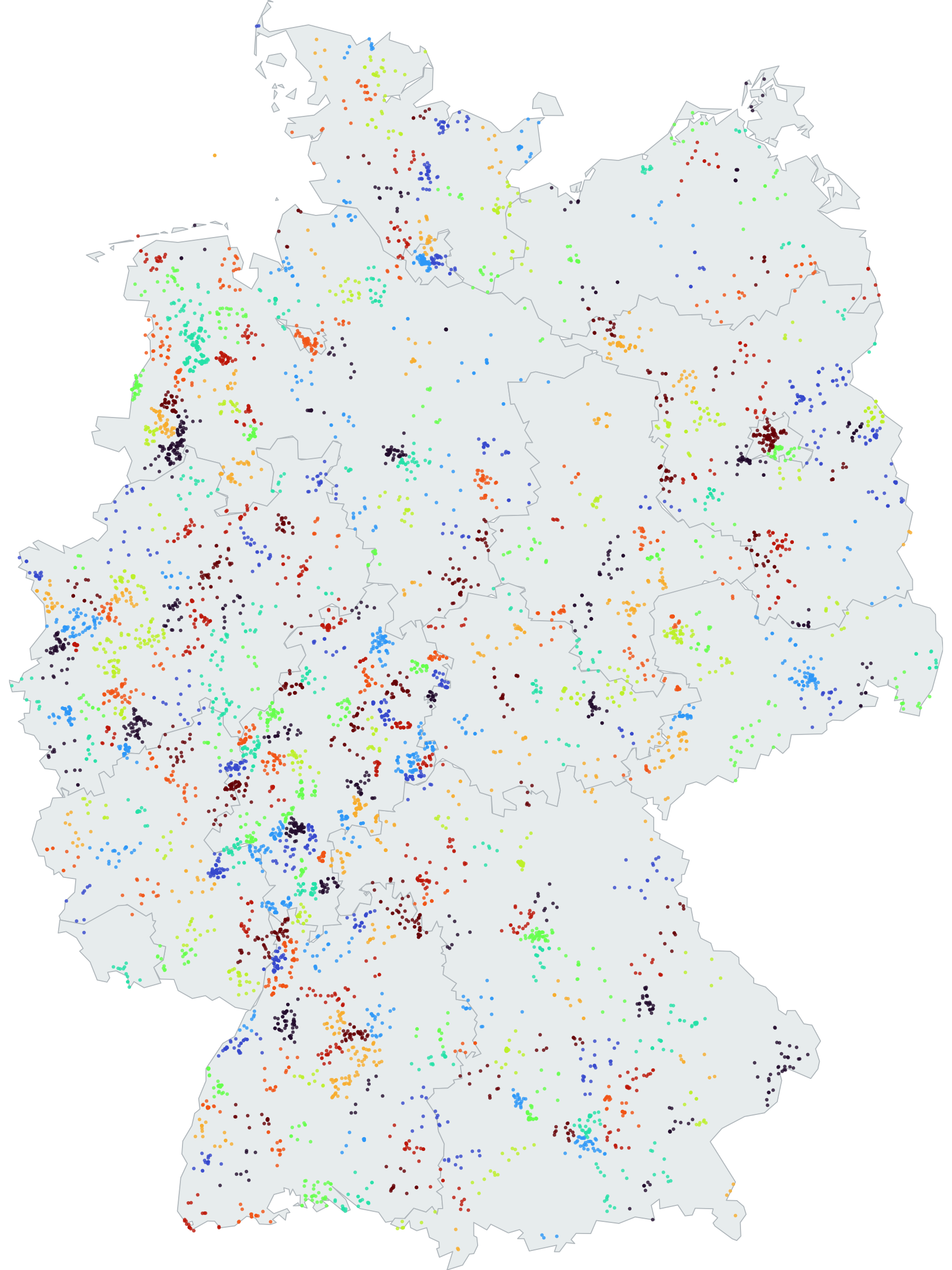}
        \caption{kNNDM Spatial Partitioning}
        \label{fig:data_overview_knndm}
    \end{subfigure}
    \caption{Overview of the study area: (\subref{fig:data_overview_samples}) spatial distribution of the training dataset across Germany and (\subref{fig:data_overview_knndm}) partitioning of the training dataset into 10 folds using the kNNDM algorithm, where each unique color represents a fold.}
    \label{fig:geographic_setup}
\end{figure}

The methodology relies on three open data sources: vector geometries from OSM to define the sampling areas, official registry data from the German MaStR to validate and clean the OSM-derived reference data, and multispectral Sentinel-2 imagery as predictor variables to classify PV systems across Germany.
We used a snapshot of Germany from November 2025 to extract labeled geometries for training and to provide the building footprints required for the final application (see \autoref{chap:application}). The geographic distribution of the resulting ground truth dataset is shown in Figure \ref{fig:data_overview_samples}.
A challenge with OSM data in this context is that many PV systems are not yet mapped. To prevent unmapped PV systems from being incorrectly included in the negative training class (rooftops without PV systems), we implemented a filtering step to identify and remove false negatives (see \autoref{fig:unified_workflow}) based on information from the MaStR. While the MaStR is often subject to registration delays and does not provide information on smaller PV systems ($\leq 30\,\mathrm{kW}$, $\sim 200\,\mathrm{m}^2$), using it to cross-reference OSM data ensures a higher level of data quality. Any building initially classified as having no PV systems installed that was within a 50-meter distance of a MaStR-registered system was removed from the dataset.

To obtain the predictor variables, we retrieved imagery from the Sentinel-2 mission via the Planetary Computer STAC API \citep{microsoft_planetary_2025}. We selected Level-2A bottom-of-atmosphere reflectance values \citep{main2015calibration, main2017sen2cor}. Our predictors consisted of the visible spectrum (B02, B03, B04), the red-edge and near infrared bands (B05, B06, B07, B08, B8A), and the short wave infrared (SWIR) bands (B11, B12).

To generate the reference data, we implemented the filtering pipeline shown in Figure \ref{fig:unified_workflow}. For the positive class (PV), we extracted OSM geometries with a minimum size of \SI{200}{\metre\squared} that were tagged as either \texttt{generator:\allowbreak source=\allowbreak solar} or \texttt{generator:\allowbreak method=\allowbreak photovoltaic}, combined with a \texttt{location} or \texttt{generator:\allowbreak location} tag equal to \texttt{roof}. For the negative class, we selected industrial building geometries tagged as \texttt{industrial}, \texttt{warehouse}, \texttt{factory}, \texttt{manufacture}, \texttt{retail}, \texttt{supermarket}, \texttt{commercial}, \texttt{logistics}, or \texttt{depot} while excluding residential categories and any structure with a footprint smaller than \SI{200}{\metre\squared}. From each footprint, we calculated the geometric centroid as location for a training sample for model training. By sampling the center, we maximized the distance to the edges, thereby reducing mixed-pixel artifacts where the reflectance values might be influenced by shadows or surrounding surfaces like asphalt or vegetation \citep{richards2006remote}. The dataset was then balanced to a 1:1 class ratio by randomly subsampling the negative class locations to match the total number of available PV samples, providing the final set of 8520 coordinates for spectral feature extraction (see \autoref{fig:data_overview_samples}).

Following the generation of sampling locations, we extracted reflectance values from the ten multispectral bands described above, resampling all bands to a 10 m resolution using nearest neighbor resampling. To use representative spectral values and minimize the effect of atmospheric conditions by filtering out spectral noise \citep{roberts2017high}, we used the median value from the three clearest (i.e, with the lowest cloud coverage) Sentinel-2 scenes for each location between January and November 2025.

To evaluate the performance of the model, we split the data into a training set (80\%) and a hold-out test set (20\%). Standard random sampling often leads to optimistic bias due to spatial autocorrelation \citep{Ploton2020, Roberts2017}. To address this, we used k-fold Nearest Neighbor Distance Matching (kNNDM) as a prediction domain adaptive sampling method to perform the split into training and test data. This method partitions the data into folds such that the geographic distances between training and test sets are similar to the distances between the training data and the actual areas being predicted \citep{linnenbrink2023knndm}. This approach creates testing conditions that reflect the challenge the model faces when mapping new, unobserved regions, i.e. the rest of Germany.

\subsection{Model training and validation}
We trained a Random Forests classifier on the 80\% training subset (i.e. 6896 samples). To determine the optimal hyperparameter configuration during model tuning, we implemented a 10-fold cross-validation using the kNNDM method. The resulting spatial partitions for the cross-validation are visualized in \autoref{fig:data_overview_knndm}. Model tuning involved a grid search to tune the number of variables selected at each split (\texttt{mtry}) where we tested values between 2 and the number of predictor variables. The number of trees was kept constant at a value of 1000. Variable importance was assessed using the permutation method via the \texttt{ranger} engine \citep{ranger}. After training, the model was applied to the 20\% test subset (i.e. 1724 samples) and the prediction was compared to the true label by calculating validation metrics including the accuracy and the area under the receiver operating characteristic (ROC) curve (AUC). 

To also assess how the framework handles regional variations in industrial architecture and OSM mapping density, we used a "leave-state-out" validation strategy. The model was iteratively trained on data from 15 federal states and tested on the single state held out. This geographic cross-validation acts as a test for model transferability. It reveals whether the spectral patterns learned in one part of Germany remain similar to regions with different building styles. This approach measures the model's ability to generalize to new geographic areas where the combination of roofing materials and environment might differ from the training set.

To provide regional prediction quality diagnostics, we calculated the dissimilarity index (DI) and the local point density (LPD) according to \citet{meyer_predicting_2021, schumacher_estimation_2024}. The DI quantifies the distance between a new observation and the training data in the multidimensional predictor space, with each predictor weighted by its respective importance in the model. Complementing this, the LPD provides additional diagnostic information by indicating the density of training data within a specific distance threshold in the feature space. Hence the results information on how well a new location resembles the training data.

\section{Results}
\label{chap:results}

\subsection{Spectral Importance and Signature}
\autoref{fig:var_imp} reveals that the SWIR bands (B11 and B12) were the most important predictors in the model, probably due to the material-specific spectral properties of PV modules (ref). The PV class had lower mean reflectance values in the visible bands (B02–B04), but this gap narrowed in the infrared range. In the SWIR region (i.e. Band 11), the reflectance for PV was higher than that of industrial roofing without PV.

\subsection{Overall Classification Performance}
\begin{figure}[htbp]
    \centering
    \begin{minipage}{0.48\textwidth}
        \centering
        \includegraphics[width=\textwidth]{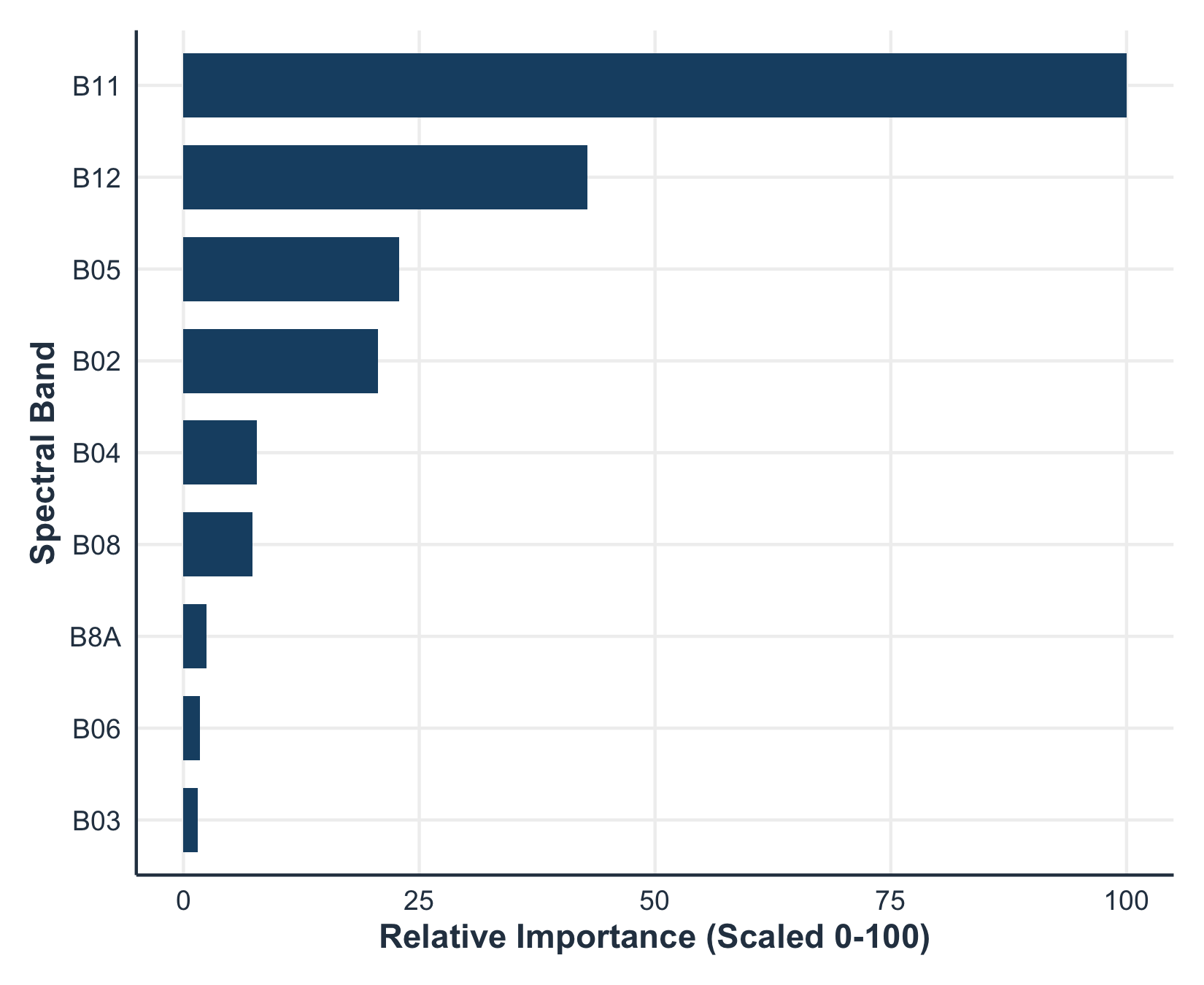}
        \caption{Feature importance.}
        \label{fig:var_imp}
    \end{minipage}
    \hfill 
    \begin{minipage}{0.48\textwidth}
        \centering
        \includegraphics[width=\textwidth]{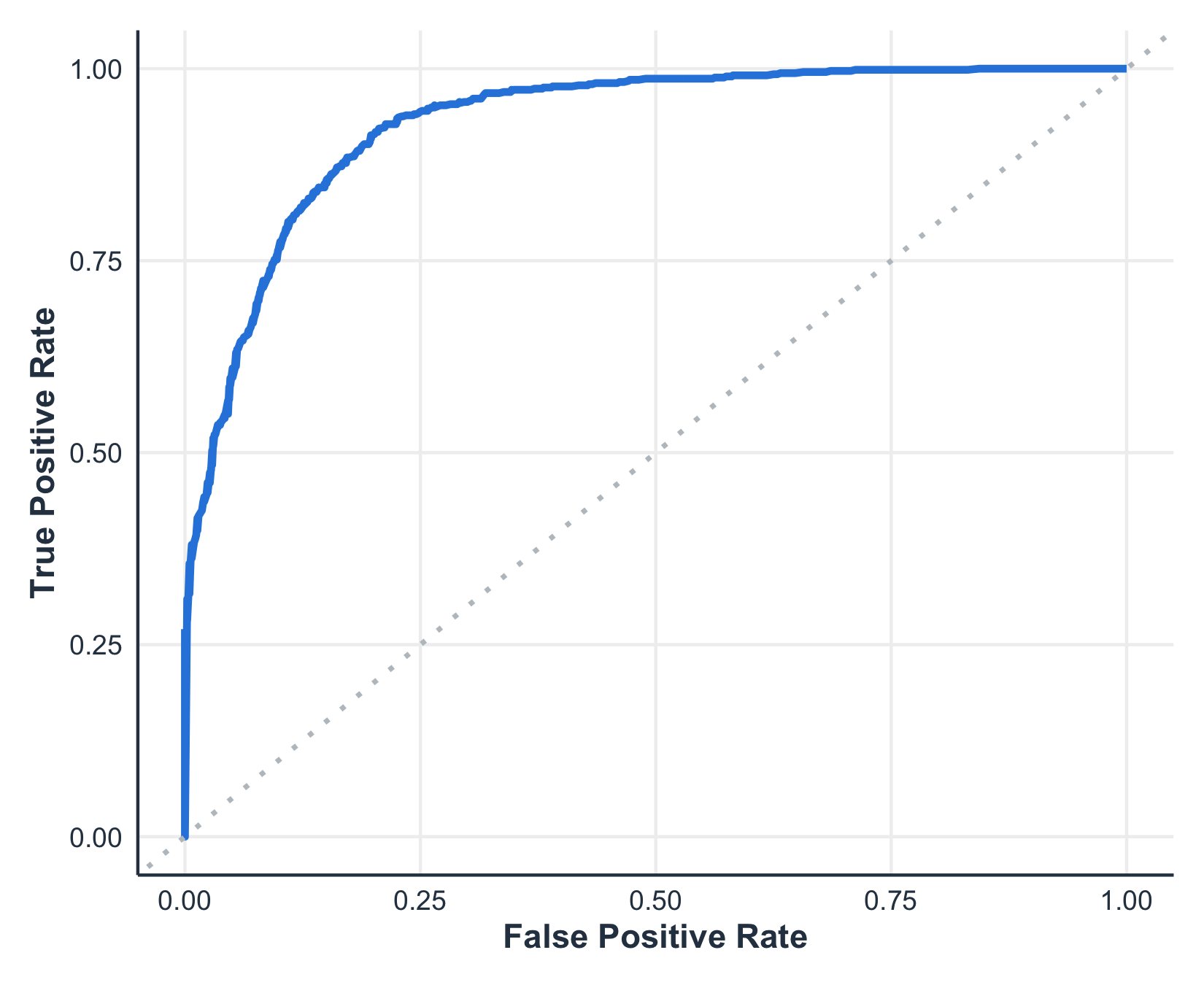}
        \caption{ROC Curve for the hold-out test set.}
        \label{fig:roc_results}
    \end{minipage}
\end{figure}

The evaluation using the 20\% hold-out test set indicated an accuracy of 85.22\% and a Kappa of 0.704 (\autoref{tab:performance_metrics}).
The framework reached a sensitivity of 0.835 and a specificity of 0.873. The calculated AUC of 0.93 (\autoref{fig:roc_results}) further indicates that these spectral features provide strong class separability across varying decision thresholds.

\begin{table}[htbp]
\centering
\caption{Model performance metrics based on the test set.}
\label{tab:performance_metrics}
\begin{tabular}{lr}
\hline
Metric & Value \\ \hline
Overall Accuracy & 85.22\% \\
Cohen's Kappa & 0.704 \\
F1-Score & 0.860 \\
Sensitivity (Recall) & 0.835 \\
Specificity & 0.873 \\
Area Under Curve (AUC) & 0.930 \\ \hline
\end{tabular}
\end{table}

\subsection{Performance by Footprint Size}
\label{sec:area_sensitivity}

The model performance, analyzed using the 20 \% test set, varied across the different footprint areas. As shown in \autoref{tab:area_results}, the accuracy recorded for rooftops smaller than 500 $m^2$ was 80.2\%. Accuracy increased alongside the footprint size, reaching 91.1\% for the largest category ($> 2500$ $m^2$). In this largest category, the model correctly classified 113 out of 124 samples.

\begin{table}[htbp]
\centering
\caption{Classification accuracy categorized by footprint area bins.}
\label{tab:area_results}
\begin{tabular}{lrr}
\hline
Area ($m^2$)      & Sample Size ($n$) & Accuracy \\ \hline
$<$ 500           & 374               & 80.2\%   \\
500--1000         & 673               & 85.6\%   \\
1000--2500        & 553               & 86.8\%   \\
$>$ 2500          & 124               & 91.1\%   \\ \hline
\end{tabular}
\end{table}

\subsection{Spatial Transferability}
\begin{figure}[htbp]
    \centering
    \includegraphics[width=0.5\textwidth]{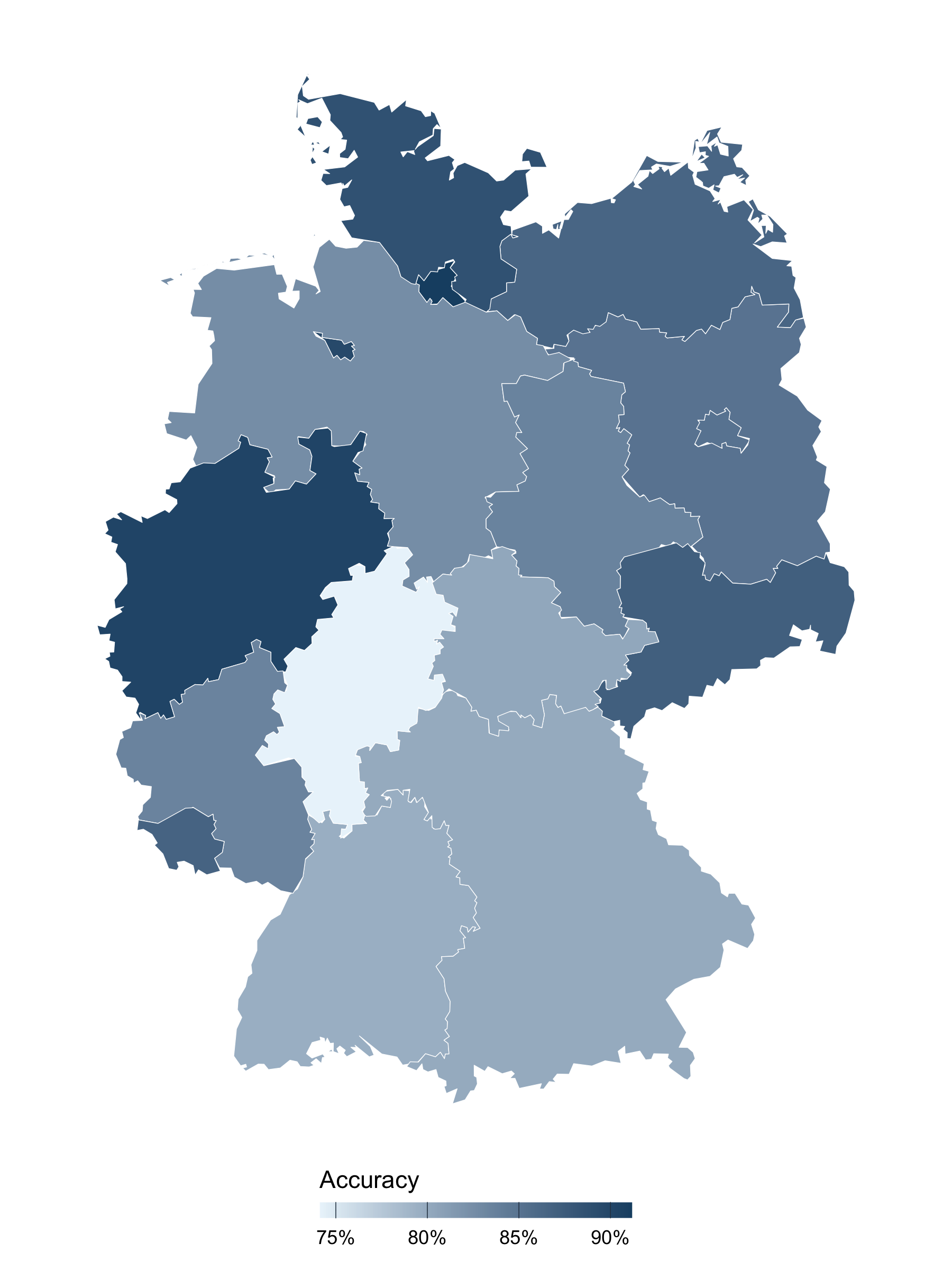}
    \caption{Leave-state-out map: Prediction accuracy for each federal state when trained on data from the remaining 15 states.}
    \label{fig:lbo_map}
\end{figure}

The leave-state-out analysis provided a measure of the model's spatial transferability across the 16 federal states of Germany. The mean accuracy was 84.61\%, close to the performance on the hold-out test set (85.22\%). Geographic variations in performance were observed across the federal states. While the model reached an accuracy of 91.2\% in Hamburg, it fell to 74.2\% in Hessen. \autoref{fig:lbo_map} visualizes the accuracy of the prediction for each state.

\section{Application Prototype}
\label{chap:application}
\begin{figure}[htbp]
    \centering
    \includegraphics[width=0.75\textwidth]{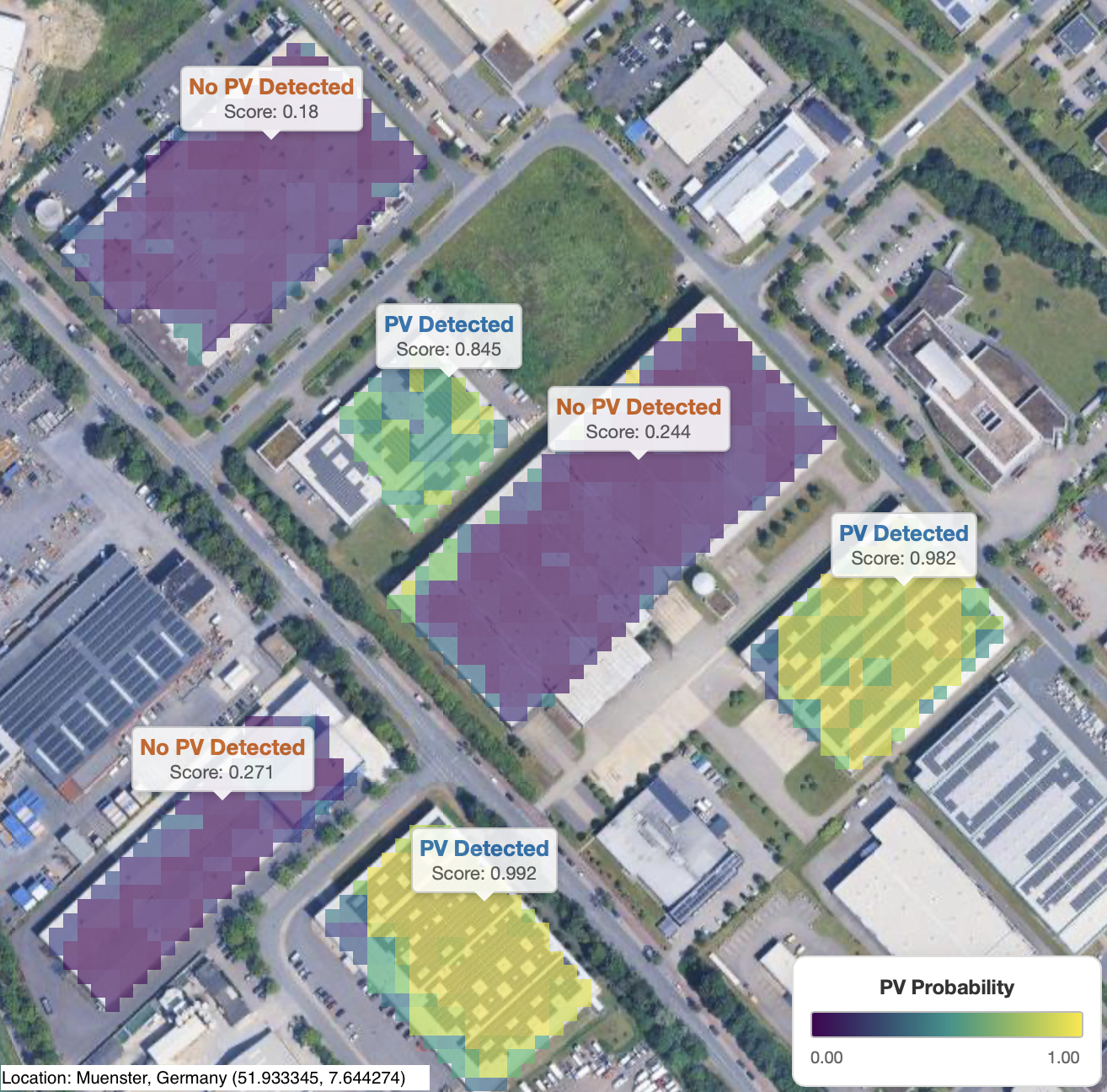}
    \caption{Interactive web interface of the prototype application. The screenshot shows the probability results for selected buildings overlaid on background imagery provided by Google Maps (\copyright Google). Green areas indicate a low probability of PV presence, while red represents a high probability.}
    \label{fig:web_interface}
\end{figure}

The model described above can be applied to new Sentinel-2 scenes across Germany. 
We developed a prototypical interactive web application designed for on-demand detection of installed PV systems (\autoref{fig:web_interface}). This application demonstrates the scalability of the framework, allowing users to inspect specific industrial sites using the most recent satellite observations rather than relying on potentially outdated commercial imagery. The prototype is published as open source software and the source code is available on GitHub\footnote{\url{https://github.com/freds-dev/pv-detection-app}}, including instructions on how to use the application.

\subsection{System Architecture}

The application uses a modular setup to coordinate data between the database, external APIs, and the prediction model. A Python backend handles the general application logic and user requests. All scientific calculations and model predictions are processed using R scripts.

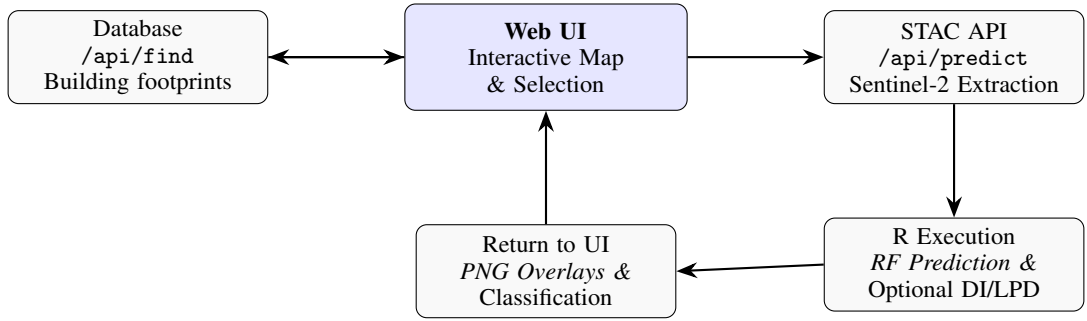
\begin{figure}[ht!]
    \centering
    \begin{tikzpicture}[
        node distance=1.2cm and 1.8cm,
        auto,
        box/.style={rectangle, draw, fill=gray!5, text width=3.2cm, align=center, minimum height=3.5em, rounded corners, font=\small},
        ui/.style={rectangle, draw, fill=blue!10, text width=3.5cm, align=center, minimum height=4em, rounded corners, font=\small\bfseries},
        line/.style={draw, -{Stealth[scale=1.2]}, thick}
    ]

        \node [ui] (ui) {Web UI \\ \textnormal{Interactive Map \& Selection}};
        
        \node [box, left=of ui] (db) {Database \\ \texttt{/api/find} \\ Building footprints};
        
        \node [box, right=of ui] (stac) {STAC API \\ \texttt{/api/predict} \\ Sentinel-2 Extraction};
        
        \node [box, below=1.5cm of stac] (script) {R Execution \\ \textit{RF Prediction \&} \\ Optional DI/LPD};
        
        \node [box, below=1.5cm of ui] (return) {Return to UI \\ \textit{PNG Overlays \&} \\ Classification};

        \draw [line] (ui.west) -- (db.east);
        \draw [line] (db.east) -- (ui.west);
        
        \draw [line] (ui.east) -- (stac.west);
        \draw [line] (stac.south) -- (script.north);
        \draw [line] (script.west) -- (return.east);
        \draw [line] (return.north) -- (ui.south);

    \end{tikzpicture}
    \caption{System architecture of the prototype application. The Flask backend orchestrates building retrieval via PostGIS and multispectral data extraction, while an external R script handles the RF inference and reliability metrics.}
    \label{fig:app_architecture}
\end{figure}

When a user clicks on the map, the system sends a request to the \texttt{/api/find} endpoint, which executes a k-Nearest Neighbor (k-NN) query using a GIST index in PostGIS to retrieve the GeoJSON geometry of the closest building. Once a building is identified, the API returns the information to the interface. The user can then select a different building or add other buildings to the current selection. Once the selection is confirmed, triggering the prediction (\texttt{/api/predict}) prompts the system to fetch the latest Sentinel-2 multispectral bands from the Microsoft Planetary Computer STAC API, filtered for cloud coverage of at or below 20\%.

Before the actual prediction, the satellite data is processed to match the building geometry. This includes resampling all spectral bands to a 10 m resolution and clipping them to the specific building footprint.
The features are then passed to an R environment via a subprocess. This script utilizes the \texttt{caret} and \texttt{CAST} packages to perform the predictions. Optionally, the quality diagnostics, i.e. the dissimilarity index (DI) and the local point density (LPD) according are calculated to inform stakeholders of whether a prediction is supported by sufficient training data.

\subsection{Interface and Visualizations}

\begin{figure}[htbp]
    \centering
    \begin{subfigure}[b]{0.32\textwidth}
        \centering
        \includegraphics[width=\textwidth]{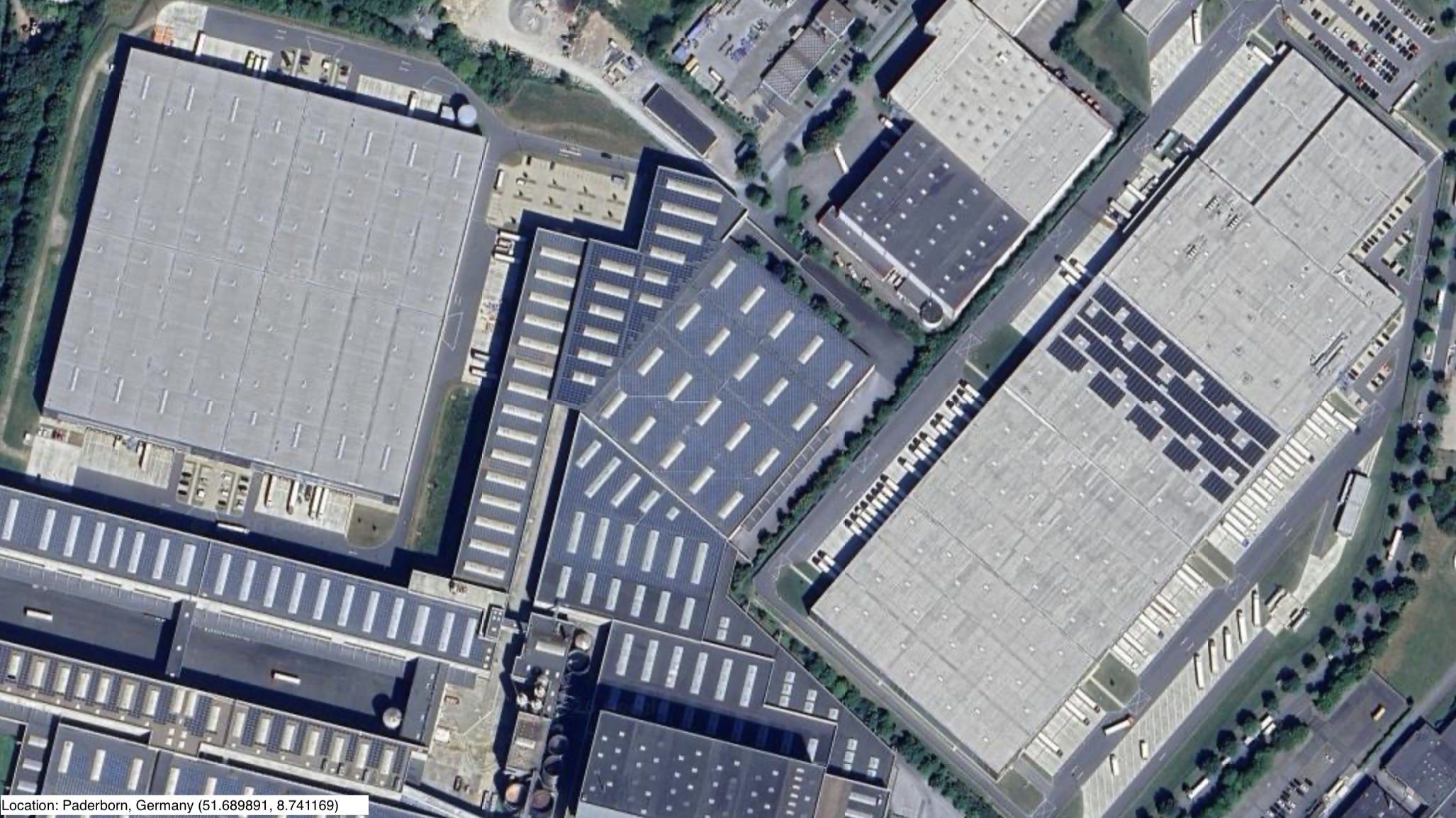}
        \caption{High-Resolution}
        \label{fig:class_rgb}
    \end{subfigure}
    \begin{subfigure}[b]{0.32\textwidth}
        \centering
        \includegraphics[width=\textwidth]{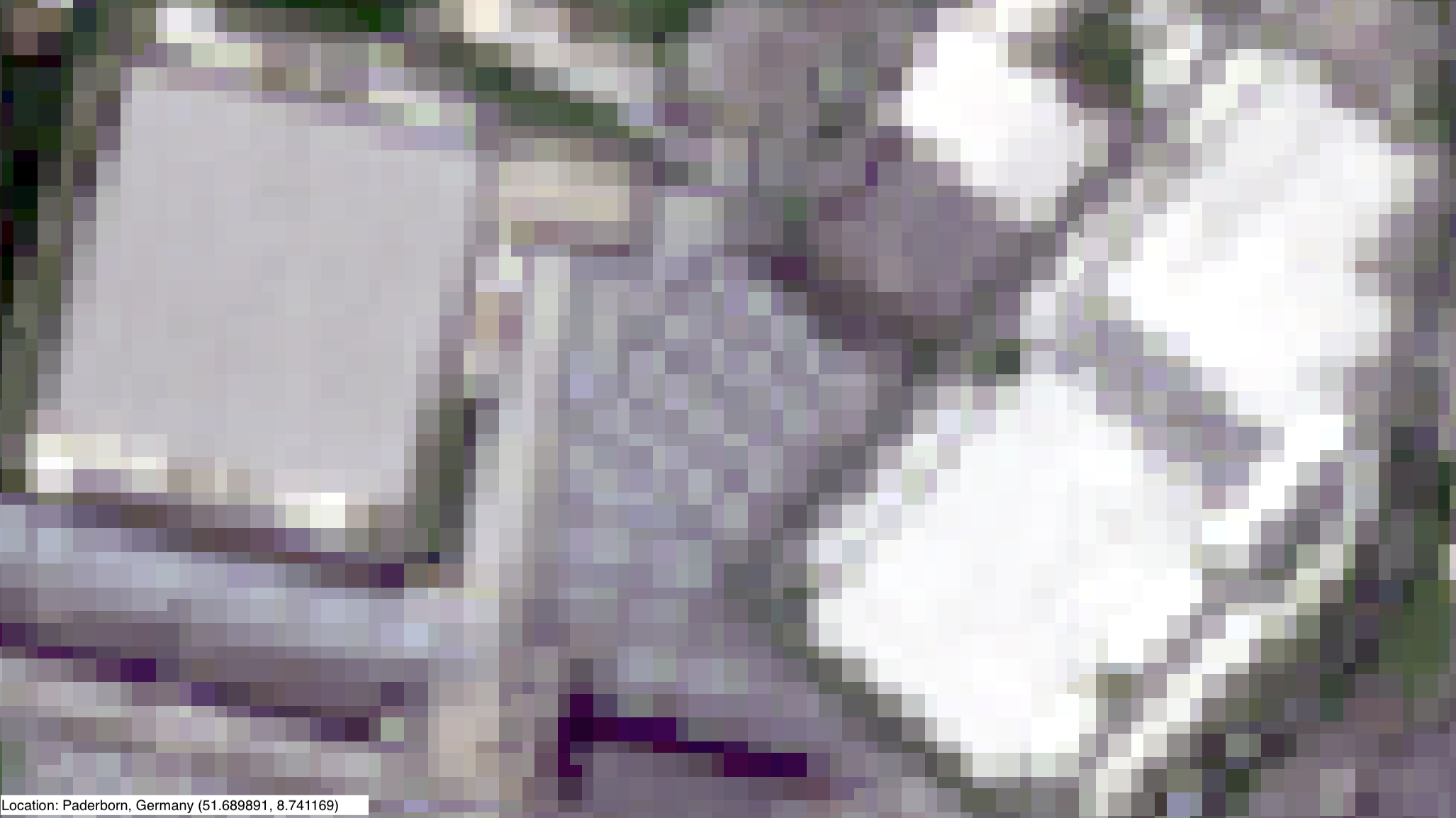}
        \caption{Sentinel-2 Overlay}
        \label{fig:class_sentinel}
    \end{subfigure}
    \hfill
    \begin{subfigure}[b]{0.32\textwidth}
        \centering
        \includegraphics[width=\textwidth]{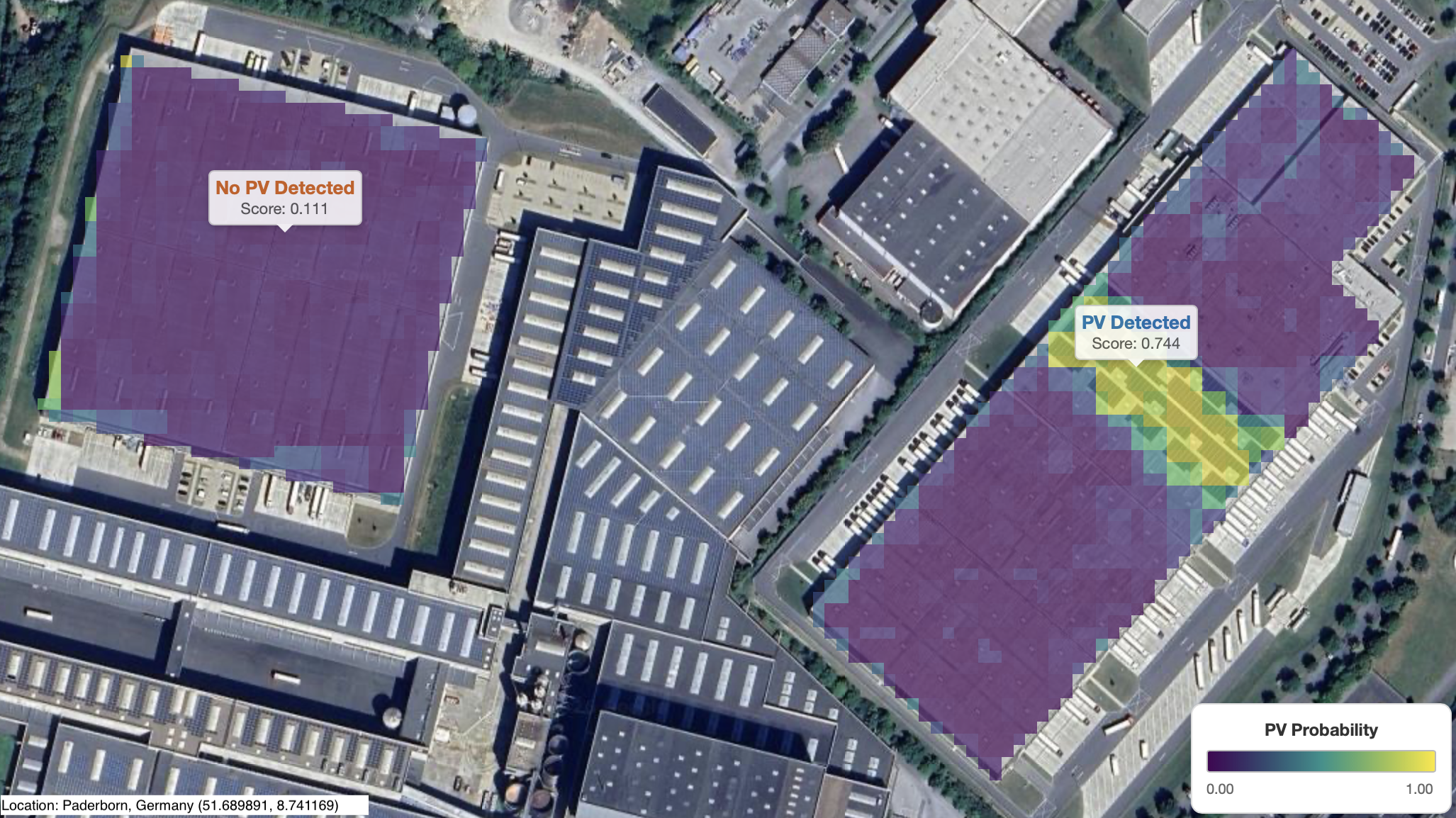}
        \caption{PV Probability}
        \label{fig:class_pred}
    \end{subfigure}
    \caption{Percentile-based classification strategy. The $P_{90}$ metric captures the presence of the array by focusing on the high-confidence pixels within the footprint. Background imagery in a and c provided by Google Maps (\copyright Google)}
    \label{fig:classification_strategy}
\end{figure}

The prediction and reliability results from the R subprocess are returned to the backend, which generates PNG overlays for the user interface, as shown in \autoref{fig:web_interface}. To translate these pixel-level probabilities into a binary classification for the entire building, we implemented an aggregation rule based on the probability scores within the building footprint. If the mean probability of the 90th percentile ($P_{90}$) exceeds a certain threshold, the building is classified as a PV site.  This threshold was determined by computing the building-level $P_{90}$ score for all buildings in the held-out test set and selecting the cutoff that resulted in the highest F1 score. This method allows for the detection of systems that occupy only a small area of the total roof area (\autoref{fig:classification_strategy}).

\begin{figure}[htbp]
    \centering
    \begin{subfigure}[t]{0.32\textwidth}
        \centering
        \includegraphics[width=\textwidth]{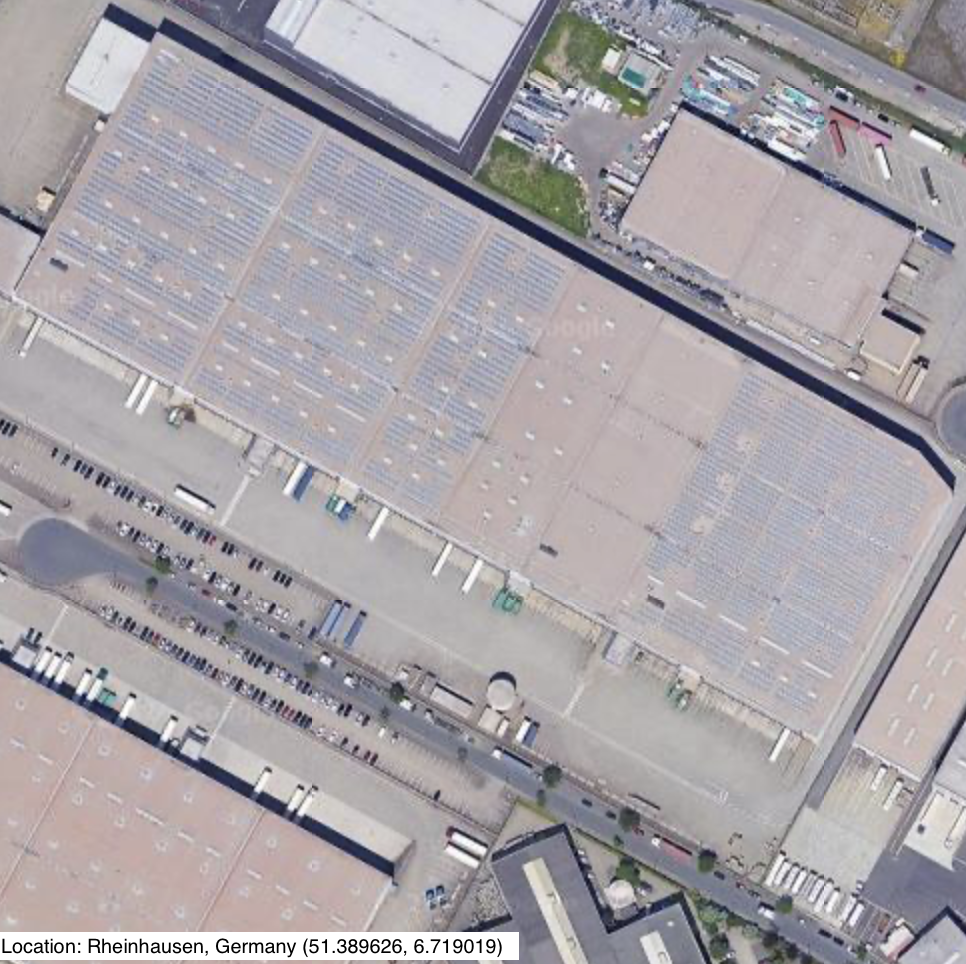}
        \caption{High-Resolution}
        \label{fig:lpd_rgb}
    \end{subfigure}
    \hfill
    \begin{subfigure}[t]{0.32\textwidth}
        \centering
        \includegraphics[width=\textwidth]{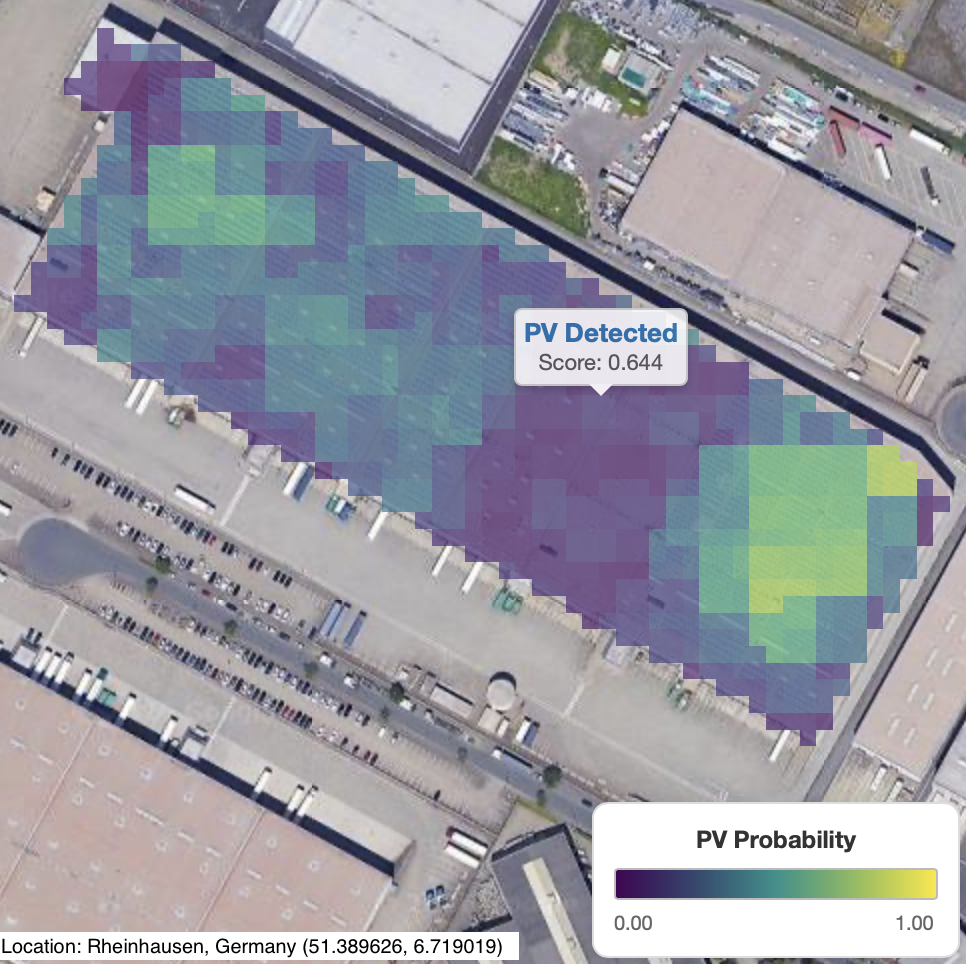}
        \caption{PV probability}
        \label{fig:lpd_pred}
    \end{subfigure}
    \hfill
    \begin{subfigure}[t]{0.32\textwidth}
        \centering
        \includegraphics[width=\textwidth]{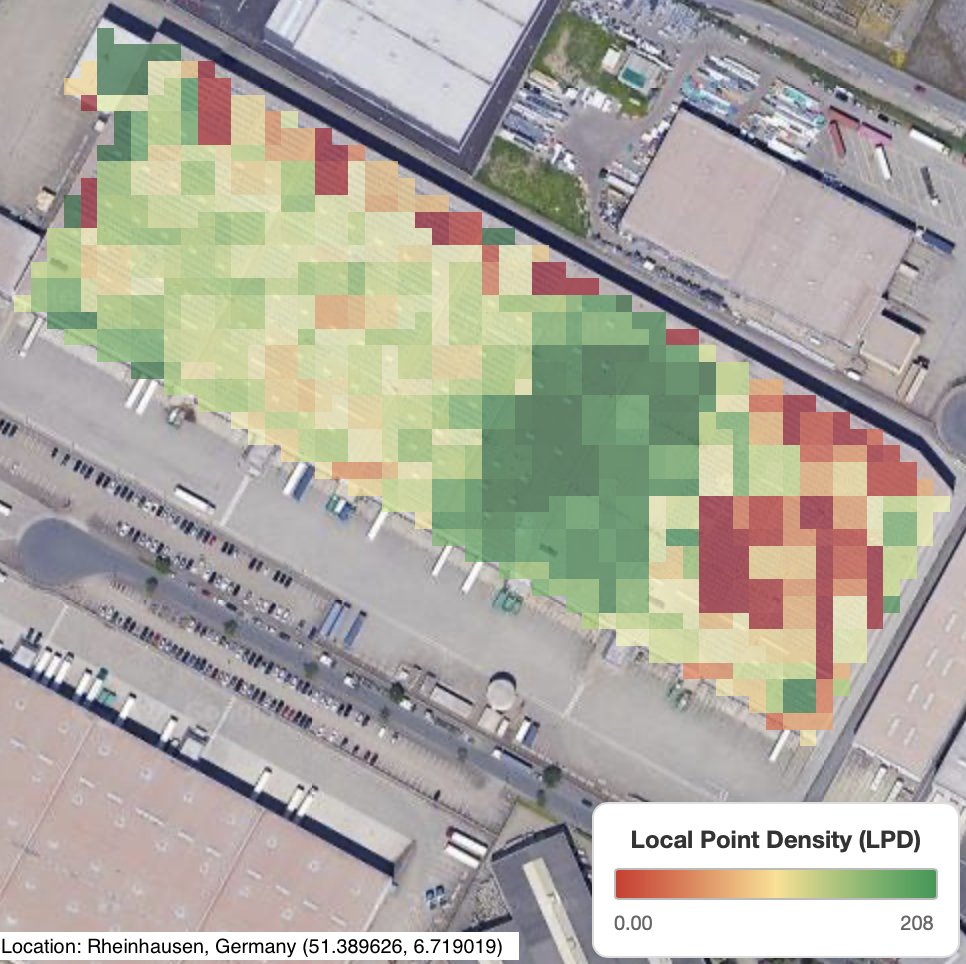}
        \caption{Local point density}
        \label{fig:lpd_layer}
    \end{subfigure}
    \caption{Visualization of the reliability assessment. The LPD overlay reveals whether the model's prediction is based on sufficient training data density in the feature space. Background imagery provided by Google Maps (\copyright Google)}
    \label{fig:lpd_demonstration}
\end{figure}

The practical application of the reliability metrics (DI, LPD) within the prototype is demonstrated in \autoref{fig:lpd_demonstration}. The high-resolution reference imagery (not used in the model; just for visual comparison) shows an industrial building with two distinct photovoltaic arrays separated by a central gap (\autoref{fig:lpd_rgb}). The corresponding model output correctly identifies the absence of PV in the central section and only returns a medium to low probability for the actual installed arrays (\autoref{fig:lpd_pred}).

A high LPD, indicated by red tones in the center of the building, confirms that the prediction for the gap is supported by a large number of similar training samples. Conversely, the red and yellow tones over the PV arrays signify a low to medium LPD, which suggests that there were fewer similar data points in the training set. By integrating these reliability metrics, the application allows users to distinguish between results with high training support and those that may require manual verification due to lower data density.

\section{Discussion}
\label{chap:discussion}

The results demonstrate that integrating Sentinel-2 multispectral imagery with a Random Forest classifier provides a scalable framework for monitoring and detection of PV infrastructure on industrial building rooftops. Achieving an accuracy of 85\% in spatially independent test regions suggests that this methodology can effectively bridge the information gap created by expensive, infrequently updated high-resolution commercial imagery. By leveraging the open-access Copernicus Sentinel-2 mission data, this framework offers a viable path for maintaining up-to-date registries without the temporal trade-offs inherent in free high-resolution datasets.

The results highlight the model’s dependence on the SWIR bands (B11 and B12), indicating that the model is driven by underlying material-specific spectral properties rather than simple visual brightness. The high reflectance of PV modules in these bands, contrasted with the absorption patterns of common roofing substrates like concrete or bitumen, is attributed to the characteristic properties of silicon cells and polymer encapsulants such as ethylene-vinyl acetate \citep{jorges2023detection, ji2021solar}. By capturing these distinct signatures, the model can differentiate PV arrays from complex urban backgrounds, even at a 10 m spatial resolution where visual identification in only the RGB spectrum is challenging.

The results demonstrate strong identification performance overall with a high accuracy of 91.1\% achieved for larger facilities, however, with some limitations for smaller structures due to the spatial resolution. The lower accuracy observed for structures under \SI{500}{\metre\squared} is a direct consequence of the 10 m spatial resolution of Sentinel-2 imagery, where small buildings are represented by a limited number of pixels. This increases the "mixed-pixel" problem, where spectral signals from the surrounding environment or roofing materials are averaged with the PV signal. However, since the majority of the industrial buildings have roofs of a larger size, a substantial share of industrial capacity can be tracked reliably using open data, reducing the need for costly commercial imagery.

There were regional variations in performance, most notably the comparably low accuracy of 74.2\% in Hessen. 
These regional variations may indicate that geographic differences in industrial architecture or roofing materials can introduce spectral noise not fully captured during training. However, this does not explain the low accuracy for Hessen where the uncertainties (reflected by the LPD values) were low, i.e. similar spectral properties were observed in the training phase. One possible explanation is that the model has a lower recall than precision. This has a stronger impact on the overall performance in areas with higher number of existing PV systems such as Hessen.
The regional variations are further complicated by inconsistent OSM data coverage between urban and rural areas \citep{ahmouda2018analyzing}. 

Although the model was developed and tested for Germany, it is expected to generalize to regions with similar roof structures or can be adapted to new areas by incorporating region-specific training data.

\section{Conclusions}
This study demonstrates the potential of using Sentinel-2 data for the timely detection of PV systems on industrial rooftops in Germany. The results indicate high classification accuracy, largely driven by distinctive reflectance patterns in the SWIR bands. While limitations remain for smaller buildings, the approach provides reliable results for the majority of industrial structures.
The model has been integrated into an open-source application that enables on-demand assessment of installed PV systems on user-defined industrial rooftops. The app provides probabilities for PV installation and a binary classification, and a complementary LPD layer which provides an additional quality diagnostic for on-demand site monitoring. This tool can support stakeholders in monitoring and facilitating the efficient expansion of photovoltaic capacity.

\section*{Declaration of Generative AI and AI-assisted Technologies}
During the preparation of this work, the authors used ChatGPT (OpenAI) and Gemini (Google) to assist with language editing, improving clarity and readability of the manuscript, and supporting software development tasks. After using these tools, the authors carefully reviewed, revised, and verified all generated content as needed and take full responsibility for the content of the published article.

\section*{Acknowledgements}
This research was partially funded in the course of TRR 391 Spatio-temporal Statistics for the
Transition of Energy and Transport (520388526) by the Deutsche Forschungsgemeinschaft (DFG,
German Research Foundation).

\newpage

\bibliographystyle{agsm}
\bibliography{references}

\end{document}